\documentclass[journal,twoside,web]{ieeecolor}
\usepackage{lcsys}
\usepackage{amsmath,amssymb,amsfonts, amsbsy}
\usepackage{mathtools}
\usepackage{bm}
\usepackage{algcompatible}
\usepackage{algorithm}
\usepackage{array}
\usepackage{paralist}
\usepackage{textcomp}
\usepackage{stfloats}
\usepackage{url}
\usepackage{verbatim}
\usepackage[caption=false]{subfig}
\usepackage{graphicx}
\usepackage{comment}
\usepackage{mathrsfs}
\usepackage{mathcomp}
\usepackage[table]{xcolor}

\usepackage{amsthm}

\usepackage[style=ieee,doi=false,isbn=false,url=false,citestyle=numeric-comp]{biblatex}
\AtEveryBibitem{
  \clearfield{note}
}
\AtEveryBibitem{\clearfield{month}}
\AtEveryCitekey{\clearfield{month}}
\AtEveryBibitem{\clearlist{language}}

\newcommand{\defeqn}{\coloneqq}

\theoremstyle{definition}

\DeclareMathOperator*{\argmax}{arg\,max}

\usepackage{color}

\usepackage{hyperref}
\hypersetup{
    colorlinks=true,
    citecolor=black,
    linkcolor=black,
    filecolor=black,      
    urlcolor=blue,
    pdfpagemode=FullScreen,
}
\usepackage[capitalize]{cleveref}	
\crefname{prb}{Problem}{Problem}

\def\BibTeX{{\rm B\kern-.05em{\sc i\kern-.025em b}\kern-.08em
    T\kern-.1667em\lower.7ex\hbox{E}\kern-.125emX}}
\begin{document}

\title{
On Generating Explanations for Reinforcement Learning Policies: An Empirical Study
}

\author{Mikihisa Yuasa$^{1}$, Huy T. Tran$^{1}$, and Ramavarapu S. Sreenivas$^{1}$
\thanks{$^{1}$The Grainger College of Engineering, University of Illinois Urbana-Champaign, Urbana, IL 61801 USA.
        {\tt\small \{myuasa2,huytran1,rsree\}@illinois.edu}}%
\thanks{*This work was supported in part by ONR grant N00014-20-1-2249 and JASSO Study Abroad Support Program (Graduate Degree Program).}
}


\maketitle
\thispagestyle{empty}

\begin{abstract}
Explaining \textit{reinforcement learning} policies is important for deploying them in real-world scenarios. We introduce a set of \textit{linear temporal logic} formulae designed to provide such explanations, and an algorithm for searching through those formulae for the one that best explains a given policy. Our key idea is to compare action distributions from the target policy with those from policies optimized for candidate explanations. This comparison provides more insight into the target policy than existing methods and avoids inference of ``catch-all'' explanations. We demonstrate our method in a simulated game of capture-the-flag, a car-parking environment, and a robot navigation task.
\end{abstract}

\section{Introduction}
\label{sec:intro}

\IEEEPARstart{R}{einforcement} learning (RL) has shown promising results for learning to autonomously make decisions. The integration of deep learning with RL, or \textit{deep reinforcement learning} (DRL), has significantly advanced the field by, for example, surpassing human experts in games \cite{silver_general_2018} and contributing to robotics \cite{ju_transferring_2022}.
However, the complexity of DRL models raises a significant concern: their lack of explainability  \cite{heuillet_explainability_2021}.
As these systems advance to solve real-world problems, their decision-making becomes more opaque, making it difficult to understand the reasoning behind their choices. Such understanding could, for example, improve trust in a self-driving car, where a passenger might like to know why a car suddenly stopped, or be used by an engineer during the design process to verify that a navigation policy satisfies the desired specification. 

A common approach to explaining complex policies, such as those produced by DRL, is to mine \textit{temporal logic} (TL) explanations from observed system behaviors \cite{bartocci_survey_2022}.
TL offers a formal structure and interpretable semantics for describing the evolution of system states over time. Traditionally, methods are based on a classification setup, where positive (from the target policy) and negative (from another policy) trajectories are provided, with various techniques used to optimize the classifier.
For example, \cite{gaglione_learning_2021,kong_temporal_2017} use directed acyclic graphs to simultaneously learn a \textit{signal temporal logic} (STL) formula structure and associated parameters, while \cite{bombara_offline_2021,karagulle_safe_2024} use tree-based algorithms. Neural networks have recently been used to learn parameters of a given \textit{weighted signal temporal logic} (wSTL) formula structure \cite{yan_stone_2021} and the structure itself \cite{li_learning_2023}. However, since these methods focus on differentiating positive and negative data, their inferred explanations strongly depend on how the given data were generated. Other methods only require positive data \cite{jha_telex_2017,jha_telex_2019}, but they remain biased by the given positive data. Non-TL methods have also been proposed to explain policies, including methods to identify important features \cite{lundberg2017,ribeiro2016}, identify important time steps \cite{guo2021bb}, learn state abstractions \cite{bewley2021}, and explore counterfactuals \cite{madumal2020}, but these methods provide limited interpretability with respect to the overall objective of the policy. Inverse RL methods can infer policy objectives \cite{arora2021a}, but they simply identify feature weights and therefore cannot model temporal objectives captured by TL.

We take a different approach where we assume access to the target policy itself, rather than being given positive and/or negative data.
The target policy could be accessible, for example, if one was optimizing a DRL policy for a complex task and wanted to gain insight into the underlying decision logic behind the policy to increase trust or support its verification in safety-critical applications.
We are motivated by the idea that having access to the target policy will provide additional information that could help with the inference process.
More specifically, we use action distributions from the target policy to gain insight into an agent's preferences across actions for sampled states.
This insight could be useful, for example, when an agent does not have a strong preference for the action taken in a given state, which could help differentiate between possible explanations for that policy.
Inferring such preferences from a fixed dataset (i.e., without access to the policy itself) is challenging, particularly for stochastic policies and environments.
These preferences could also be used to emphasize informative states during the inference process, which may not have been visited in a given set of trajectories.

Based on this idea, we propose a search method that infers the \emph{linear temporal logic} (LTL) formula that best explains a target policy, where this explanation delineates the operational conditions maintained throughout execution and the ultimate objectives achieved.
Our approach centers on a class of LTL formulae that is endowed with a concept of neighborhood that is amenable to a local-search (\Cref{subsec:node,subsec:nbh}). 
Each potential LTL-explanation is then translated into an RL policy, which is compared to the target policy using a well-structured metric (\Cref{subsec:node_eval}). 
Should a neighboring explanation have better alignment with the target policy, it supplants the current explanation. 
This iterative process persists until none of the neighboring explanations surpasses the present one, thus establishing a local optimum as the recommended explanation.
To enhance robustness, we propose additional neighborhood expansion and extension heuristics (\Cref{subsec:exp_ext}), as well as a multi-start implementation of the search that generates the top-$k$ candidate explanations.
We demonstrate our method in three simulated environments (\Cref{sec:results}).

\section{Background}
\label{sec:background}

\subsection{Reinforcement Learning}
\label{subsec:mdp}

We model our problem as a \textit{Markov decision process} (MDP) $\mathcal{M} \defeqn \langle \mathcal{S}, \mathcal{A}, p, r, \gamma \rangle$ \cite{sutton_between_1999}, where $\mathcal{S}$ is the state space, $\mathcal{A}$ is the action space, and $\gamma \in [0, 1)$ is the discount factor. Then $p(s'|s,a): \mathcal{S} \times \mathcal{A} \times \mathcal{S} \mapsto [0,1]$ is the state transition function and $r(s,a,s'): \mathcal{S} \times \mathcal{A} \times \mathcal{S} \mapsto \mathbb{R}$ is the reward function for states $s,s' \in \mathcal{S}$ and action $a \in \mathcal{A}$. Let $\pi(a|s): \mathcal{S} \times \mathcal{A} \mapsto [0,1]$ be a policy for the agent. At time step $t$, the agent executes action $a_t \sim \pi(\cdot | s_t)$ given the current state $s_t$, after which the system transitions to state $s_{t+1}$ and the agent receives reward $r(s_t, a_t, s_{t+1}$). The objective is to learn a policy that maximizes $\mathbb{E}_{p,\pi} \left[ \sum_{t=0}^\infty \gamma^{t} r(s_t, a_t, s_{t+1}) \mid s_0 = s, a_0 = a \right]$.

\subsection{Linear Temporal Logic and FSPA-Augmented MDPs}
\label{subsec:LTL}

The syntax for LTL is defined as $\phi\defeqn\top \bigm| \psi(s) \bigm| \neg \phi \bigm| \phi \vee \phi' \bigm| \phi \wedge \phi' \bigm| \mathcal{G}(\phi) \bigm| \mathcal{F}(\phi) \bigm| \mathcal{U}(\phi) \bigm| \mathcal{X}(\phi) $ for logical formulae $\phi$ and $\phi'$. Here, $\top$ is the True Boolean constant, $s\in\mathcal{S}$ is an MDP state, $\psi(s) \defeqn f(s) < c$ is an atomic predicate for $c\in\mathbb{R}$, and $\neg$ (negation), $\wedge$ (conjunction), and $\vee$ (disjunction) are Boolean connectives. $\mathcal{F}$ (eventually), $\mathcal{G}$ (globally), $\mathcal{U}$ (until), and $\mathcal{X}$ (next) are temporal operators.
Given an LTL formula and a state or trajectory, we can assign a real-number value, the robustness, that reflects the degree to which the formula is satisfied. More formally, given a state $s \in \mathcal{S}$ and LTL formulae $\phi, \phi'$, the robustness of boolean operators is defined as,
\begin{align*}
    &\rho(s, f(s) < c) &= & \quad c - f(s),\\
    &\rho(s, \neg\phi) &=& \quad -\rho(s, \phi),\\
    &\rho(s, \phi \wedge \phi') &=& \quad \min(\rho(s, \phi),\rho(s, \phi')),\\
    &\rho(s, \phi \vee \phi') &=& \quad \max(\rho(s, \phi),\rho(s, \phi')).
\end{align*}
Given a trajectory of states $\tau \defeqn (s_0, s_1, \dots, s_k)$, the robustness of temporal operators is defined as,
\begin{align*}
    &\rho(\tau, \mathcal{F}(\phi)) &=& \quad \max_{i\in[0,k]}(\rho(s_i,\phi)),\\
    &\rho(\tau, \mathcal{G}(\phi)) &=& \quad \min_{i\in[0,k]}(\rho(s_i,\phi)).
\end{align*}

A feasible LTL formula can be transformed into a \textit{finite state predicate automaton} (FSPA) using $\omega$-automaton manipulation \cite{schneider_improving_2001}.
An FSPA $\mathscr{A}$ is defined by $\langle \mathcal{Q}, \mathcal{S}, \mathcal{E}, \Phi, q_0, b, F, Tr \rangle$, where $\mathcal{Q}$ is a finite set of automaton states, $\mathcal{S}$ is an MDP state space, $\mathcal{E} \subseteq \mathcal{Q}\times\mathcal{Q}$ is the set of edges (transitions) between automaton states, $\Phi$ is the input alphabet, $q_0 \in \mathcal{Q}$ is the initial automaton state, $b:\mathcal{E} \mapsto \Phi$ maps an edge $(q_t, q_{t+1})$ to its transition condition (defined as an LTL formula $\phi$), $F \subseteq \mathcal{Q}$ is the set of final automaton states, and $Tr \subseteq \mathcal{Q}$ is the set of automaton trap states.

We now define an FSPA-augmented MDP $\mathcal{M}_\mathscr{A}$ based on \cite{li_formal_2019}. 
Given MDP $\mathcal{M}$ and FSPA $\mathscr{A}$, an FSPA-augmented MDP $\mathcal{M}_\mathscr{A}$ is defined by a tuple $\langle \Tilde{\mathcal{S}}, \mathcal{Q}, \mathcal{A}, \Tilde{p}, \Tilde{r}, \mathcal{E}, \Psi, q_0, b, F, Tr \rangle$, where $\Tilde{\mathcal{S}}\subseteq \mathcal{S} \times \mathcal{Q}$ is the product state space. Then $\Tilde{p}(\Tilde{s}'|\Tilde{s},a): \Tilde{\mathcal{S}} \times \mathcal{A} \times \Tilde{\mathcal{S}}  \mapsto [0,1]$ is the state transition function and $\Tilde{r}(\Tilde{s},a,\Tilde{s}'): \Tilde{\mathcal{S}} \times \mathcal{A} \times \Tilde{\mathcal{S}} \mapsto \mathbb{R}$ is the reward function for states $\Tilde{s},\Tilde{s}' \in \Tilde{\mathcal{S}}$ and action $a \in \mathcal{A}$. We consider two rewards in this paper. For sparse reward settings, we define the reward $\Tilde{r}$ as,
\begin{equation}
    \Tilde{r}(\Tilde{s},a,\Tilde{s}') = 
    \begin{cases}
        0 & q = q',\\
        -\rho(s, b(q, q')) & q' \in Tr ,\\
        \rho(s, b(q, q')) & \mathrm{otherwise},
    \end{cases}
    \label{eqn:r_tilde}
\end{equation}
where $q, q' \in \mathcal{Q}$.
For dense reward settings, we define the reward $\Tilde{r}$ as,
\begin{equation}
    \Tilde{r}(\Tilde{s},a,\Tilde{s}') = 
    \begin{cases}
        \beta\rho(s, b(q, q^*)) & q = q',\\
        -\rho(s, b(q, q')) & q' \in Tr ,\\
        \rho(s, b(q, q')) & \mathrm{otherwise},
    \end{cases}
    \label{eqn:r_tilde_dense}
\end{equation}
where $\beta\in [0, 1)$ is a scaling factor, $q^* = \argmax_{q'' \in \mathcal{N}} \rho(s,b(q,q''))$, and $\mathcal{N} = \{q'' \in \mathcal{Q}\setminus Tr | (q,q'') \in \mathcal{E} \}$ is the set of non-trap automaton states that neighbors $q$.
Intuitively, this dense reward encourages a transition to a non-trap automaton state if the next automaton state is the same as the current one.

\section{Method}
\label{sec:method}

\begin{figure}[tb]
    \centering
    \medskip
    \includegraphics[width=\linewidth]{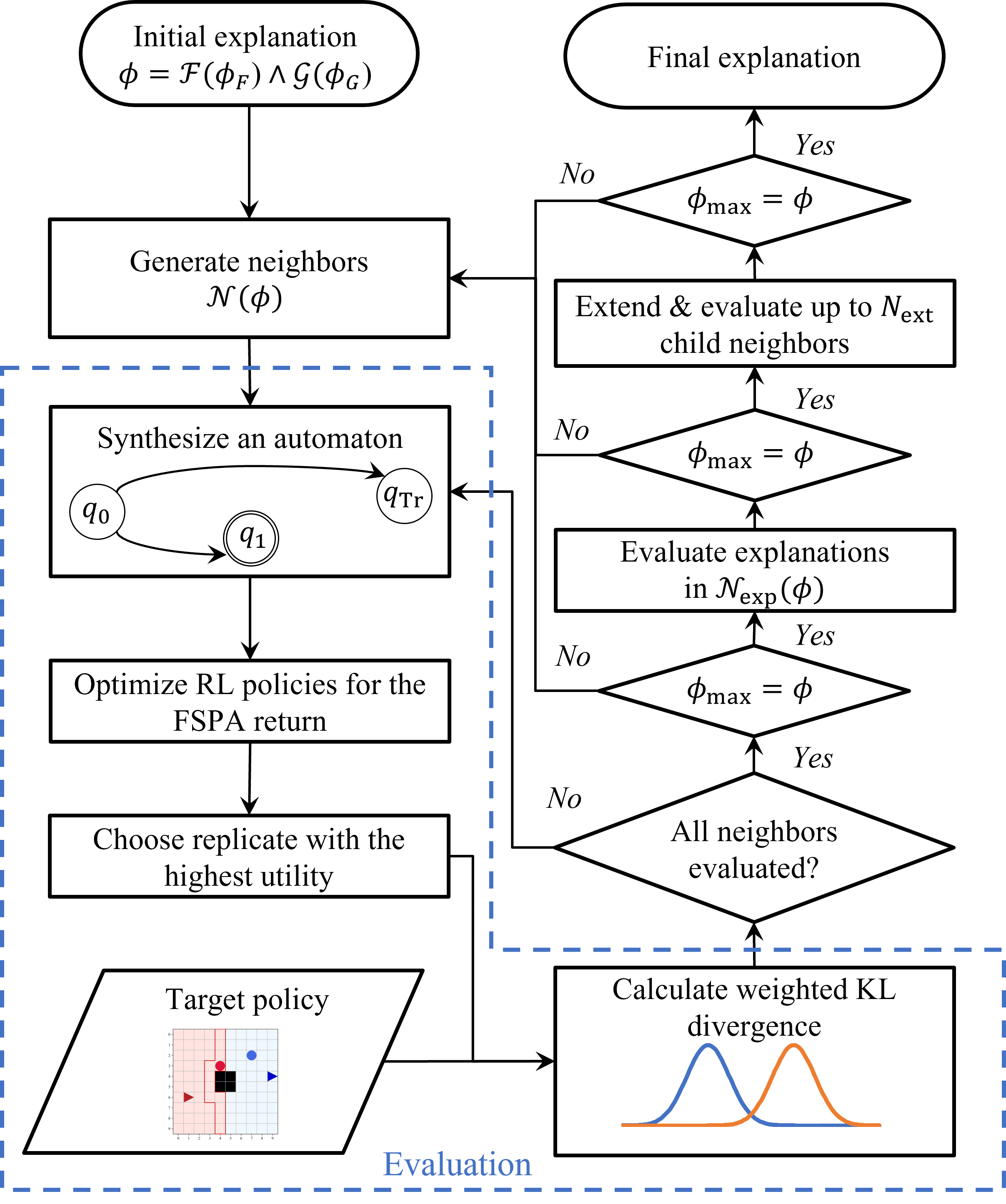}
    \caption{Overview of our proposed search algorithm.}
    \label{fig:architecture}
\end{figure}

Our problem statement is as follows: given a target policy $\pi_\mathrm{tar}$, MDP $\mathcal{M}$, and a set of atomic predicates $\Psi$, find an explanation $\phi$ that explains the objectives of the policy and the operational conditions maintained throughout execution.
We address this problem by proposing a greedy local-search method, summarized in \Cref{fig:architecture} and \Cref{alg:search,alg:neighbors}. 
We detail the key components of our method below.

\subsection{Definition of Explanations}
\label{subsec:node}
We consider explanations of the form $\phi = \mathcal{F}(\phi_F) \wedge \mathcal{G}(\phi_G)$, where $\phi_F$ and $\phi_G$ are \textit{Conjunctive Normal Form} (CNF) or \textit{Disjunctive Normal Form} (DNF) formulae that have up to two clauses within them.
That is, we assume the policy tries to achieve a task represented by $\phi_F$ while satisfying global constraints represented by $\phi_G$. 
For example, given atomic predicates $\Psi \defeqn \{\psi_0, \psi_1, \psi_2, \psi_3, \psi_4\}$, a possible explanation is $\mathcal{F}(\psi_{0} \vee\psi_{1}) \wedge \mathcal{G}(\neg\psi_{2} \wedge (\neg\psi_{3} \vee \psi_{4}))$, which requires that: ``\textit{Eventually, either $\psi_0$ or $\psi_1$ is satisfied. Globally, $\psi_2$ is not satisfied and either $\psi_3$ is not satisfied or $\psi_4$ is satisfied}.''

We represent each explanation as a row vector of truth values whose length is $3N_\mathrm{pred}+2$, where $N_\mathrm{pred} = |\Psi|$. 
The first $N_\mathrm{pred}$ elements define whether or not each predicate is negated (0 for no negation, 1 for negation). 
The next $N_\mathrm{pred}$ elements define which temporal formula, $\phi_F$ or $\phi_G$, each predicate belongs to (0 for $\phi_F$, 1 for $\phi_G$). 
We require each temporal formula to contain at least one predicate. 
The following $N_\mathrm{pred}$ elements define which clause within the temporal formula each predicate belongs to (0 for the first clause, 1 for the second clause).
The last two elements define whether each temporal formula, $\phi_F$ or $\phi_G$, is in CNF or DNF form (0 for CNF, 1 for DNF).

\subsection{Evaluation of Explanations}
\label{subsec:node_eval}

We evaluate the utility of a candidate explanation by measuring the similarity between the target policy and a policy optimized for that LTL formula.
A naive evaluation could simply generate a set of trajectories from the target policy and then count the trajectories that satisfy a formula.
However, this approach would give high utility to ``catch-all'' explanations (e.g., those containing disjunctions of all predicates).
We instead propose to measure utility as the weighted average of Kullback-Leibler (wKL) divergence values between the action distributions of the target policy and the policy optimized for a candidate explanation over sampled states.
More specifically, we first synthesize an FSPA-augmented MDP $\mathcal{M}_\mathscr{A}$ using a candidate explanation $\phi$ and use RL to optimize a policy with respect to reward $\Tilde{r}$ (using \Cref{eqn:r_tilde} or \Cref{eqn:r_tilde_dense}), following the method proposed by \cite{li_formal_2019}.
We address the fact that multiple optimal policies can exist by optimizing $N_\mathrm{rep}$ replicates and choosing the policy with the highest policy entropy $H^\pi$, calculated as,
\begin{equation}
    H^\pi = \frac{-\sum_{s\in \mathcal{B}_\mathrm{NT}} A_\pi(s) \log A_\pi(s)}{|\mathcal{B}_\mathrm{NT}|},
    \label{eqn:entropy}
\end{equation}
where $A_\pi(s) \defeqn \pi(\cdot|s)$ is the action distribution for policy $\pi$ at state $s$ and $\mathcal{B}_\mathrm{NT}$ is a set of randomly sampled non-trap states (i.e., states $s \in \mathcal{S}$ whose corresponding automaton state is in $\mathcal{Q}\setminus Tr$).

We then measure the similarity between the selected policy and the target policy by calculating KL divergence values over action distributions of sampled states. More specifically, for a selected policy $\pi_\phi$, target policy $\pi_\mathrm{tar}$, and state $s$, we calculate,
\begin{align}
\label{eqn:kl_divergence}
    D_\mathrm{KL}(A_{\pi_\phi}(s) || A_{\pi_\mathrm{tar}}(s)) &= A_{\pi_\phi}(s) \log \frac{A_{\pi_\phi}(s)}{A_{\pi_\mathrm{tar}}(s)}.
\end{align}
If policies output multiple actions, the average of KL divergence values over action distribution pairs is used.
Finally, we use \Cref{eqn:kl_divergence} to calculate the utility $U$ of explanation $\phi$ as a wKL divergence value over states $s_i \in \mathcal{B}_\mathrm{NT}$,
\begin{align}
\label{eqn:weighted_kl_divergence}
    U^\phi = -\sum_{i = 1}^{|\mathcal{B}_\mathrm{NT}|}w_i D_\mathrm{KL}( A_{\pi_\phi}(s_i) || A_{\pi_\mathrm{tar}}(s_i)),
\end{align}
where $w_i$ is the weight associated with $s_i$, calculated as,
\begin{align}
\label{eqn:weight}
    w_i = \frac{\Bar{H}^{\pi_\mathrm{tar}}(s_i)}{\sum_{s_j \in \mathcal{B}_\mathrm{NT}} \Bar{H}^{\pi_\mathrm{tar}}(s_j)}.
\end{align}
Here, $\Bar{H}^{\pi_\mathrm{tar}}(s)$ is a normalized entropy calculated as,
\begin{align}
\label{eqn:normalized_entropy}
    \Bar{H}^{\pi_\mathrm{tar}}(s) = 1 + \frac{A_{\pi_\mathrm{tar}}(s)\log A_{\pi_\mathrm{tar}}(s)}{H_\mathrm{max}},
\end{align}
where $H_\mathrm{max}$ is the maximum possible entropy. 
We use a weighted average to emphasize similarity between policies at states where the target policy is highly certain about what action to take.
We include an augmented return filter to ignore any explanations that produce a policy with a converged return lower than a user-defined threshold $\Tilde{R}_\mathrm{th}$. 
This filter thus ignores explanations that are impossible to satisfy given the FSPA-augmented MDP.

\subsection{Neighborhood Definition and Evaluation}
\label{subsec:nbh}
After evaluating an explanation $\phi$, our search proceeds by generating a neighborhood $\mathcal{N}(\phi)$ of related explanations. We define the neighborhood $\mathcal{N}(\phi)$ as the set of explanations whose row vector representations differ from that of $\phi$ at a single location (i.e., a single bit-flip). The class of LTL formulae we consider (i.e., of the form $\mathcal{F}(\phi_F)\wedge\mathcal{G}(\phi_G)$) are thus completely connected under this notion of a neighborhood. 
After generating $\mathcal{N}(\phi)$, we evaluate each of the neighboring explanations using the method discussed in \Cref{subsec:node_eval}.

\subsection{Additional Neighborhood Expansion \& Extension}
\label{subsec:exp_ext}
To address the potential existence of multiple undesired local optima, we include additional neighborhood expansion and extension steps in our search. The expansion step creates an expanded neighborhood $\mathcal{N}_\mathrm{exp}(\phi)$ for a given parent explanation $\phi$ by flipping the values of the last two elements (i.e., the elements that define whether the temporal formulae are in CNF or DNF) of each explanation in the original neighborhood $\mathcal{N}(\phi)$. 
These additional explanations are then evaluated. 
This step is implemented if the neighborhood $\mathcal{N}(\phi)$ does not produce a better explanation than $\phi$.

The extension step forces the search to generate and evaluate neighborhoods for the child explanations in $\mathcal{N}(\phi)$, even when the explanations in $\mathcal{N}(\phi)$ have a lower utility than $\phi$. That is, we extend the search by evaluating neighbors of the explanations in $\mathcal{N}(\phi)$ with the highest utilities, up to $N_\mathrm{ext}$ times. This step is implemented if the original and extended neighborhoods of $\phi$ do not produce a better explanation than $\phi$.

\medskip
\begin{algorithm}[tb]
    \caption{TL Greedy Local-Search}
    \textbf{Input}: number of searches $N_\mathrm{search}$, number of maximum search steps $N_\mathrm{max}$, number of replicates $N_\mathrm{rep}$, number of sampled episodes $N_\mathrm{ep}$, return filter threshold $\Tilde{R}_\mathrm{th}$, number of extension steps $N_\mathrm{ext}$\\
    \textbf{Output}: explanation $\phi$
    \begin{algorithmic}[1]
        \STATE Construct empty buffer $\Phi$
        \STATE \textbf{for} $n = 1,2,...,N_\mathrm{search}$ \textbf{do}
        \STATE \hspace{0.25cm} Initialize starting explanation $\phi$ randomly
        \STATE \hspace{0.25cm} \textbf{for} $m = 1, 2,..., N_\mathrm{max}$ \textbf{do}
        \STATE \hspace{0.5cm} $B =\mathrm{EvalNeighbors}(\phi, N_\mathrm{rep}, N_\mathrm{ep}, \Tilde{R}_\mathrm{th})$
        \STATE \hspace{0.5cm} $\phi_{\mathrm{max}}, U_{\mathrm{max}} = B[0]$
        \STATE \hspace{0.5cm} \textbf{if} $\phi = \phi_\mathrm{max}$ \textbf{then}
        \STATE \hspace{0.75cm} \textbf{for} $i = 1, 2, ... , N_\mathrm{ext}$ \textbf{do}
        \STATE \hspace{1.0cm} $\phi', U' \leftarrow B[i]$
        \STATE \hspace{1.0cm} $B' = \mathrm{EvalNeighbors}(\phi', N_\mathrm{rep}, N_\mathrm{ep},\Tilde{R}_\mathrm{th})$
        \STATE \hspace{1.0cm} $\phi'_\mathrm{max}, U'_\mathrm{max} \leftarrow B'[0]$
        \STATE \hspace{1.0cm} \textbf{if} $U'_\mathrm{max} > U_\mathrm{max}$ \textbf{then}
        \STATE \hspace{1.25cm} $\phi, U \leftarrow \phi'_\mathrm{max}, U'_\mathrm{max}$
        \STATE \hspace{1.25cm} \textbf{break}
        \STATE \hspace{0.75cm} \textbf{if} $\phi = \phi_\mathrm{max}$ \textbf{then} \textbf{break}
        \STATE \hspace{0.5cm} \textbf{else} $\phi, U \leftarrow \phi_\mathrm{max}, U_\mathrm{max}$
        \STATE \hspace{0.25cm} Store $(\phi, U)$ to $\Phi$
        \STATE \textbf{return} $\phi$ from $\Phi$ with the highest $U$
    \end{algorithmic}
    \label{alg:search}
\end{algorithm}

\medskip
\begin{algorithm}[tb]
    \caption{EvalNeighbors()}
    \textbf{Input}: starting explanation $\phi_\mathrm{in}$, number of replicates $N_\mathrm{rep}$,  number of sampled episodes $N_\mathrm{ep}$, reward filter threshold $\Tilde{R}_\mathrm{th}$\\
    \textbf{Output}: Sorted buffer $B=[(\phi_i, U^{\phi_i})]$
    \begin{algorithmic}[1]
        \STATE Generate $\mathcal{N}(\phi_\mathrm{in})$ and construct sorted buffer $B$ 
        \STATE \textbf{for} $\phi_i$ in $\mathcal{N}(\phi_\mathrm{in})$ \textbf{do}
        \STATE \hspace{0.25cm} Synthesize automaton from $\phi_i$
        \STATE \hspace{0.25cm} Optimize and evaluate $N_\mathrm{rep}$ policies for $\phi_i$
        \STATE \hspace{0.25cm} Choose the policy with highest $U^{\phi_i}$
        \STATE \hspace{0.25cm} Calculate average return $\Bar{R}$ for $N_\mathrm{ep}$ episodes
        \STATE \hspace{0.25cm} \textbf{if} $\Bar{R} > \Tilde{R}_\mathrm{th}$  \textbf{then} store $(\phi_i, U^{\phi_i})$ to $B$
        \STATE $\phi_{\mathrm{max}}, U_{\mathrm{max}} = B[0]$
        \STATE \textbf{if} $\phi_{\mathrm{max}} = \phi_\mathrm{in}$ \textbf{then} repeat Lines 2-7 for $\mathcal{N}_\mathrm{exp}(\phi_\mathrm{in})$
        \STATE \textbf{return} $B$
    \end{algorithmic}
    \label{alg:neighbors}
\end{algorithm}

\section{Results}
\label{sec:results}

\subsection{Test Environments and RL Training Details}

\begin{figure}[tb]
    \centering
    \medskip
    \begin{tabular}{@{}c@{}}
        \subfloat[CtF environment.\label{fig:game_field}]{\includegraphics[width=0.49\linewidth]{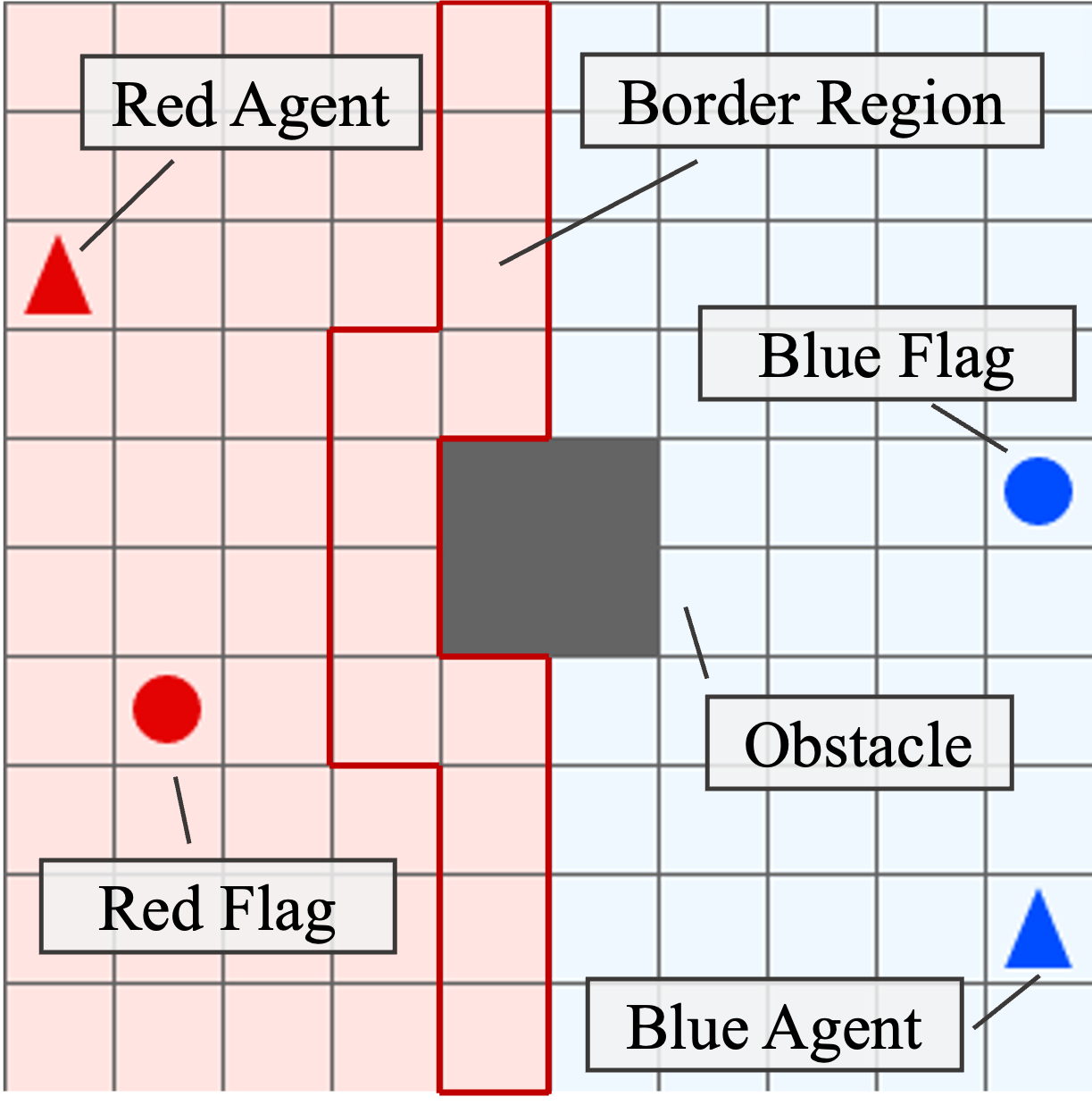}}
    \end{tabular}
    \hfill
    \begin{tabular}{@{}c@{}}
        \subfloat[Parking environment.\label{fig:parking_env}]{\includegraphics[width=0.49\linewidth]{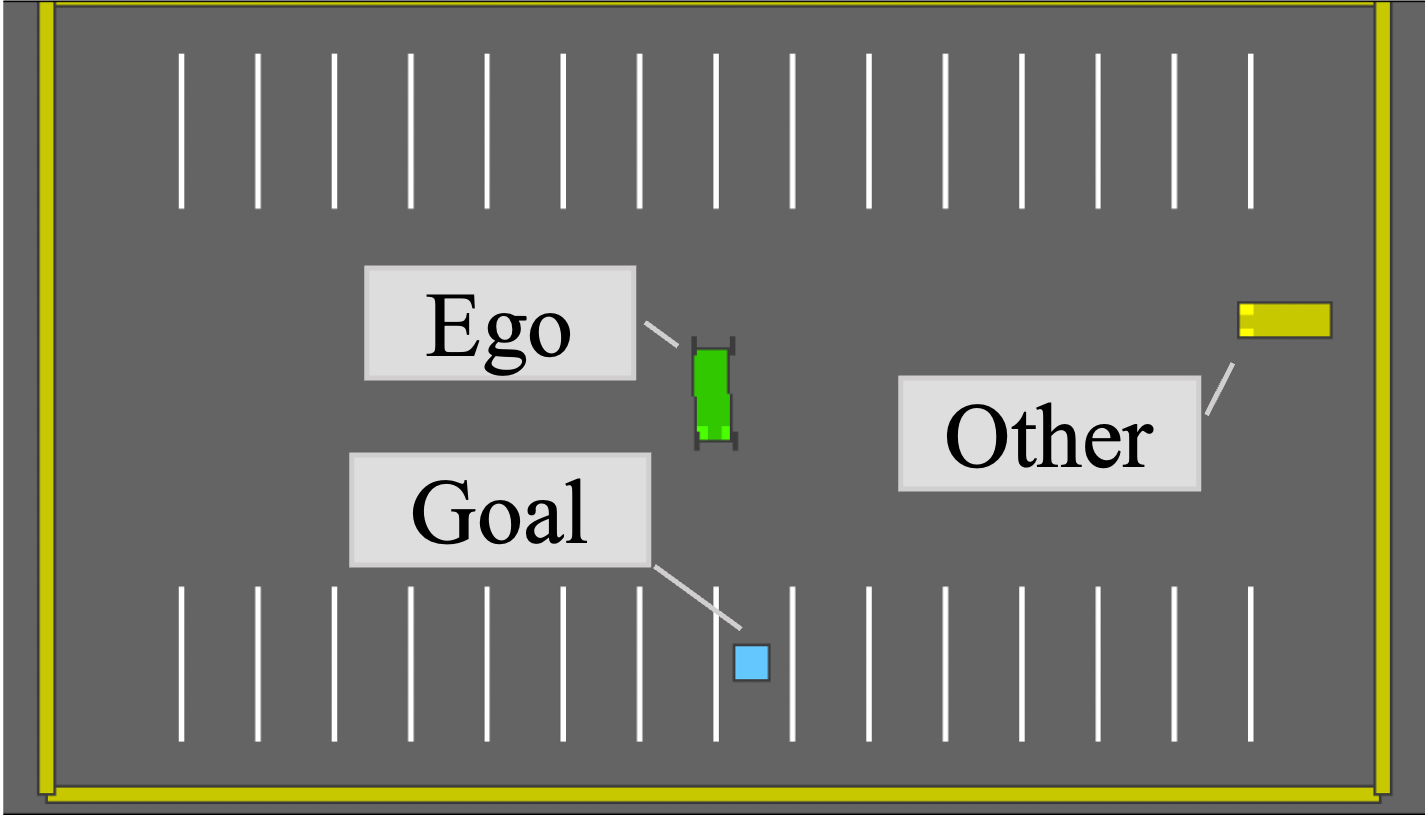}}\\
        \subfloat[Robot navigation.\label{fig:robot_env}]{\includegraphics[width=0.49\linewidth]{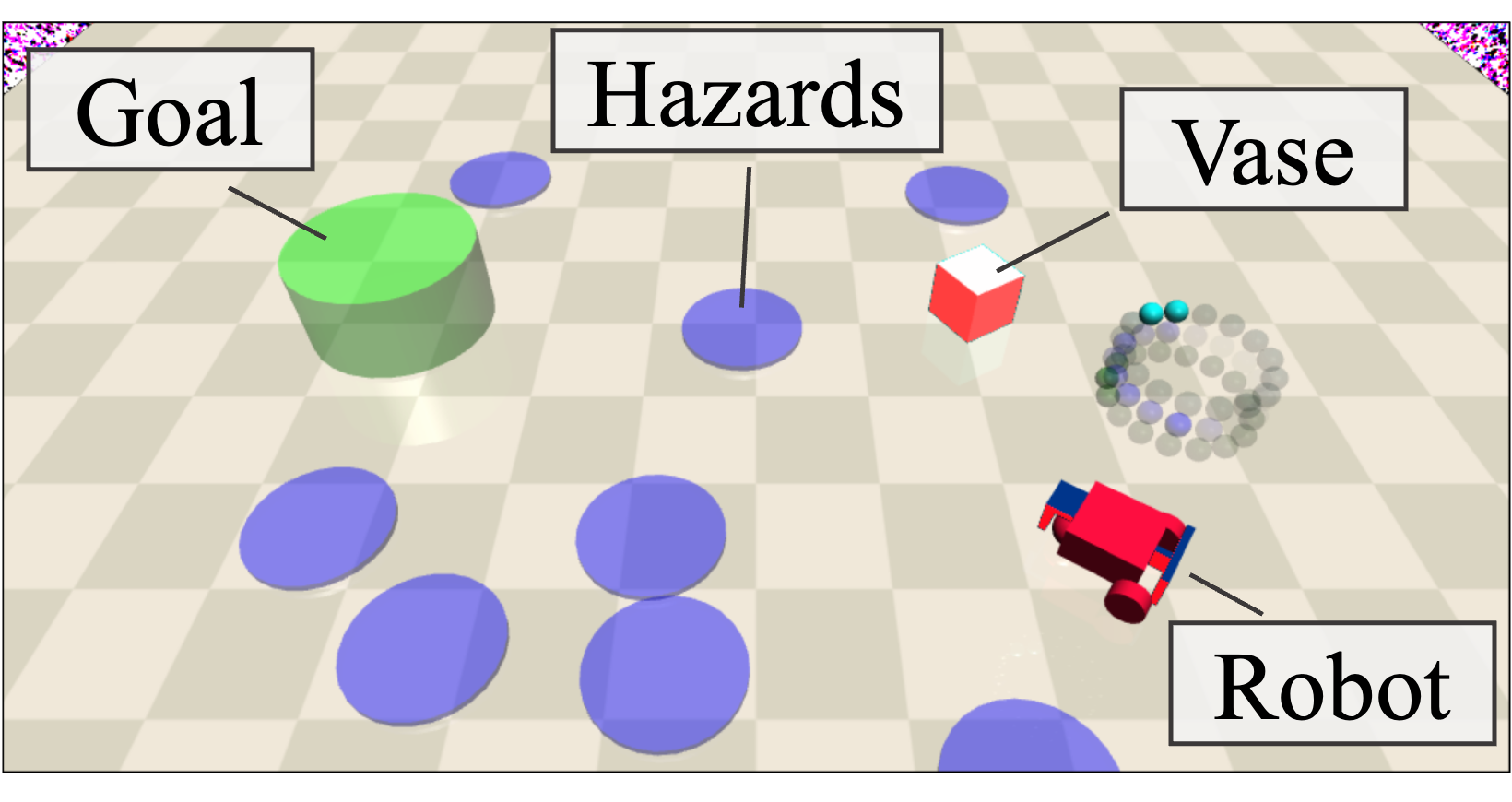}}
    \end{tabular}
    \caption{Screenshots from our environments.}
    \label{fig:envs}
\end{figure}

We considered three test environments, shown in \Cref{fig:envs}.
Capture-the-flag (CtF) is a discrete grid-world with adversarial dynamics, where agents try to capture their opponent's flag. 
If the agents are next to each other in the blue territory, the red agent is killed with 75\% probability (vice versa in the red territory).
We explained the blue agent policy and defined the red policy as a heuristic focused on defending its border region.
We defined four atomic predicates based on euclidean distances between the blue agent (ba), blue flag (bf), blue territory (bt), red agent (ra), and red flag (rf):
$\psi_\mathrm{ra, bf} = d_\mathrm{ra, bf} < 1, \psi_\mathrm{ba, rf} = d_\mathrm{ba, rf} < 1, \psi_\mathrm{ba, ra} = d_\mathrm{ba, ra} < 1.5, \psi_\mathrm{ba, bt} = d_\mathrm{ba, bt} < 1$.
The resulting search space contained 640 explanations.
Using the procedure from \cite{li_formal_2019} and sparse reward defined in \Cref{eqn:r_tilde}, we optimized the the target policy to satisfy $\mathcal{F}( \psi_\mathrm{ba, rf} \wedge \neg\psi_\mathrm{ra, bf}) \wedge \mathcal{G}(\neg\psi_\mathrm{ba, ra} \vee \psi_\mathrm{ba, bt})$, which requires that: ``\textit{Eventually, the blue agent reaches the red flag and the red agent does not reach the blue flag. Globally, the blue agent does not encounter the red agent or it stays in the blue territory.}''

We used the car-parking scenario to consider continuous state-action spaces ($s \in \mathbb{R}^{19}$, $a \in \mathbb{R}^2$), based on a modified version of highway-env \cite{highway-env}.
An episode ends when the ego vehicle reaches the landmark, hits a wall, or hits the other vehicle.
We explained the ego vehicle policy and defined the other vehicle to use a heuristic policy that moves left or right based on a random initial location.
We defined three atomic predicates: $\psi_\mathrm{g} = d_\mathrm{ego, goal} < 1, \psi_\mathrm{o} = d_\mathrm{ego, other} < 3, \psi_\mathrm{w} = d_\mathrm{ego, wall} < 4$, where ego and other stand for the ego and other vehicles, respectively.
The resulting search space contained 96 explanations.
Using the dense reward defined in \Cref{eqn:r_tilde_dense}, we optimized the target policy to satisfy $\mathcal{F}(\psi_\mathrm{g}) \wedge \mathcal{G}(\neg \psi_\mathrm{o} \wedge \neg \psi_\mathrm{w})$, which requires that: ``\textit{Eventually, the ego vehicle reaches the goal. Globally, the ego vehicle does not hit the other vehicle and walls.}''

We used a robot navigation task to consider more realistic and complex robot dynamics ($s \in \mathbb{R}^{72}$, $a \in \mathbb{R}^2$), based on a modified version of CarGoalv0 \cite{ji2023safety}.
We explained the robot policy and defined three atomic predicates: $\psi_\mathrm{gl} = d_\mathrm{robot, goal} < 0.3$, $\psi_\mathrm{hz} = d_\mathrm{robot, hazards} < 0.2$, $\psi_\mathrm{vs} = d_\mathrm{robot, vase} < 0.1$.
We considered two target policies: one optimized for an LTL specification and one optimized for a non-LTL reward (to consider a target policy that was not optimized for one of the candidate explanations).
Using the dense reward defined in \Cref{eqn:r_tilde_dense}, we optimized the LTL target policy to satisfy $\mathcal{F}(\psi_\mathrm{gl}) \wedge \mathcal{G}(\neg\psi_\mathrm{hz} \wedge \neg\psi_\mathrm{vs})$, which requires that: ``\textit{Eventually, the robot reaches the goal. Globally, the robot does not enter a hazard or vase.}''
We optimized the non-LTL policy using a reward based on distance to the goal and reaching the goal or a hazard.

We used StableBaselines3 \cite{stable-baselines3} implementations of PPO \cite{schulman_proximal_2017} to optimize policies for CtF and robot navigation, and SAC \cite{haarnoja_soft_2018} with HER \cite{andrychowicz_hindsight_2017} for parking. Hyperparameters were default ones or those suggested by RL Baselines3 Zoo \cite{rl-zoo3}. We used the following search hyperparameters, based on a small trial-and-error process: $N_\mathrm{max} = 10$, $N_\mathrm{rep} = 3 \text{ (CtF, parking) or } 1 \text{ (navigation)}$, $N_\mathrm{ep} = 200$, $\Tilde{R}_\mathrm{th} = 0.05$, and $N_\mathrm{ext} = 3$. Our code is available online\footnote{\url{https://github.com/miki-yuasa/tl-search}}.

\begin{table*}[tb]
    \centering
    \medskip
    \caption{CtF \& parking results. Target policies were successfully found in Searches 1 and 2 (CtF) and 1, 2, and 3 (parking).}
    \tabcolsep = 3pt
    \rowcolors{2}{gray!25}{white}
    \begin{tabular}{c|lcc|lcc}
         Search & CtF explanations & wKL div. [-]  & \begin{tabular}{c} Searched\\ specs [\%] \end{tabular} & Parking explanations &  wKL div. [-] & \begin{tabular}{c} Searched\\ specs [\%] \end{tabular}  \\
         \hline
         1      & $\pmb{\mathcal{F}(\psi_\mathrm{ba, rf}\wedge\neg \psi_\mathrm{ra, bf}) \wedge \mathcal{G}(\neg \psi_\mathrm{ba, ra} \vee \psi_\mathrm{ba, bt})}$                & $\mathbf{8.00\times10^{-8}}$  & \textbf{8.13} & $\pmb{\mathcal{F}(\psi_\mathrm{g}) \wedge \mathcal{G}(\neg \psi_\mathrm{o}\wedge \neg\psi_\mathrm{w})}$   & $\mathbf{0.00}$       & \textbf{37.5}  \\
         2      & $\pmb{\mathcal{F}(\psi_\mathrm{ba, rf}\wedge\neg \psi_\mathrm{ra, bf}) \wedge \mathcal{G}(\neg \psi_\mathrm{ba, ra}\vee\psi_\mathrm{ba, bt})}$                  & $\mathbf{8.00\times10^{-8}}$  & \textbf{10.2} & $\pmb{\mathcal{F}(\psi_\mathrm{g}) \wedge \mathcal{G}(\neg \psi_\mathrm{o}\wedge \neg\psi_\mathrm{w})}$   & $\mathbf{0.00}$       & \textbf{29.2} \\
         3      & $\mathcal{F}(\neg \psi_\mathrm{ba, bt}) \wedge \mathcal{G}((\neg \psi_\mathrm{ba, ra} \wedge \psi_\mathrm{ba, rf})\vee(\neg \psi_\mathrm{ra, bf}))$       & $1.46\times10^{-7}$           & 6.56 & $\pmb{\mathcal{F}(\psi_\mathrm{g}) \wedge \mathcal{G}(\neg \psi_\mathrm{o}\wedge \neg\psi_\mathrm{w})}$   & $\mathbf{0.00}$       & \textbf{37.5}  \\
         4      & $\mathcal{F}(\psi_\mathrm{ba, rf}) \wedge \mathcal{G}((\neg \psi_\mathrm{ba, bt}\wedge\neg \psi_\mathrm{ra, bf})\vee(\neg \psi_\mathrm{ba, ra}))$         & $7.42\times10^{-7}$           & 8.44 & $\mathcal{F}(\psi_\mathrm{o}) \wedge \mathcal{G}(\neg \psi_\mathrm{g}\vee \neg\psi_\mathrm{w})$     & $4.61\times10^{-4}$   & 44.8  \\
         5      & $\mathcal{F}(\neg \psi_\mathrm{ba, bt}\wedge\neg \psi_\mathrm{ra, bf}) \wedge \mathcal{G}((\neg \psi_\mathrm{ba, rf})\vee(\neg \psi_\mathrm{ba, ra}))$    & $1.57\times10^{-6}$           & 6.56 & $\mathcal{F}(\psi_\mathrm{o}) \wedge \mathcal{G}(\neg \psi_\mathrm{g}\vee \neg\psi_\mathrm{w})$     & $4.61\times10^{-4}$   & 33.3  \\
         6      & $\mathcal{F}((\neg \psi_\mathrm{ra, bf})\vee(\psi_\mathrm{ba, rf})) \wedge \mathcal{G}(\neg \psi_\mathrm{ba, ra}\vee \psi_\mathrm{ba, bt})$               & $5.95\times10^{-6}$           & 9.06 & $\mathcal{F}(\psi_\mathrm{w}) \wedge \mathcal{G}(\neg\psi_\mathrm{g} \vee \psi_\mathrm{o})$         & $6.40\times10^{-4}$   & 30.2  \\
         7      & $\mathcal{F}(\psi_\mathrm{ba, ra}) \wedge \mathcal{G}((\neg \psi_\mathrm{ba, bt})\vee(\neg \psi_\mathrm{ba, rf}\wedge\neg \psi_\mathrm{ra, bf}))$         & $1.18\times10^{-5}$           & 8.59 & $\mathcal{F}(\neg\psi_\mathrm{o}\wedge \psi_\mathrm{w}) \wedge \mathcal{G}(\neg \psi_\mathrm{g})$    & $7.35\times10^{-4}$   & 28.1  \\
         8      & $\mathcal{F}((\psi_\mathrm{ba, ra})\wedge(\neg \psi_\mathrm{ba, rf})) \wedge \mathcal{G}((\psi_\mathrm{ba, bt})\vee(\neg \psi_\mathrm{ra, bf}))$          & $1.18\times10^{-5}$           & 11.4 & $\mathcal{F}(\neg\psi_\mathrm{o}\wedge \psi_\mathrm{w}) \wedge \mathcal{G}(\neg \psi_\mathrm{g})$   & $7.35\times10^{-4}$   & 27.1  \\
         9     & $\mathcal{F}((\psi_\mathrm{ra, bf})\vee(\neg \psi_\mathrm{ba, bt})) \wedge \mathcal{G}((\neg \psi_\mathrm{ba, ra})\vee(\psi_\mathrm{ba, rf}))$             & $1.62\times10^{-5}$           & 7.19 & - & - & - \\               
         10      & $\mathcal{F}((\neg \psi_\mathrm{ba, ra})\vee(\neg \psi_\mathrm{ra, bf})) \wedge \mathcal{G}((\psi_\mathrm{ba, bt})\vee(\psi_\mathrm{ba, rf}))$           & $8.52\times10^{-5}$           & 6.56 & - & - & - \\
    \end{tabular}
    \label{tab:ctf_park_search}
\end{table*}

\subsection{Experiment Results}

\begin{figure}[tb]
    \centering
    \medskip
    \includegraphics[width=0.5\textwidth]{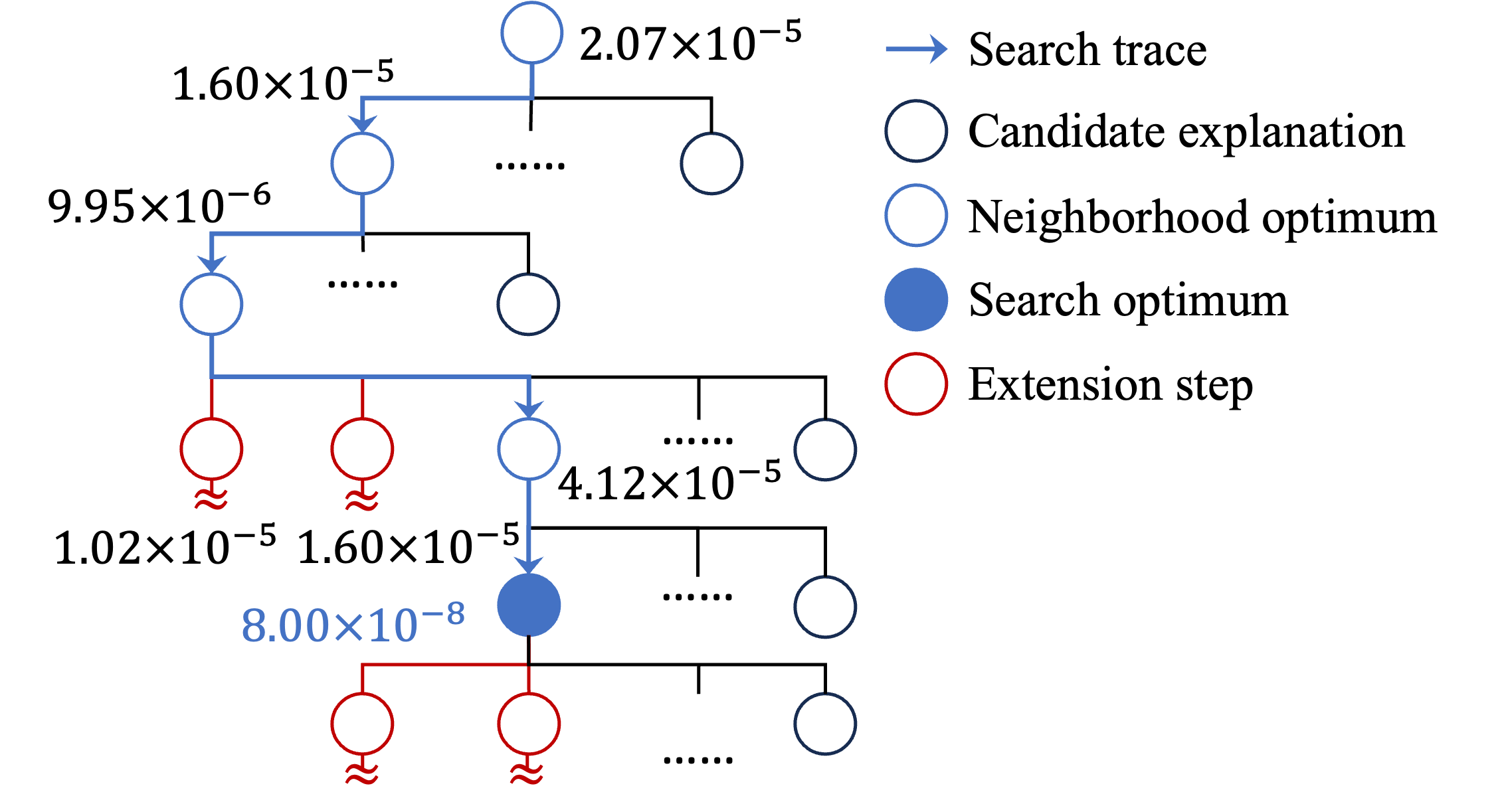}
    \caption{A partial trace of CtF Search 5 with wKL divergence values.}
    \label{fig:trace}
\end{figure}

\Cref{tab:ctf_park_search} shows our CtF results, where Searches 1 and 2 successfully found the target explanation as the one with the highest utility among searched explanations. 
The second best result was Search 6, whose explanation is: ``\textit{Eventually, the blue agent captures the red flag. Globally, the blue agent is not inside the blue territory while the red agent does not capture the blue flag, or the blue agent does not encounter the red agent.}'' 
This explanation is close to the target explanation and consistent with the game dynamics since the task includes the overall objective of capturing the red flag, while the constraint plausibly captures the battle dynamics of CtF which could incentivize the blue agent to avoid the red agent if it is outside of the blue territory, but engage the red agent within the blue territory.
\Cref{fig:trace} shows a partial trace of the search tree produced by Search 2. 
We see that the extension step was used in the third step of the search to avoid an undesired local minimum early in the process.
Though not shown, the extension step was also used in this search.

\Cref{tab:ctf_park_search} also shows our parking results, where Searches 1-3 successfully found the target explanation. 
The second best results were Searches 4 and 5, whose explanation is:
``\textit{Eventually, the ego vehicle encounters the other agent. Globally, the ego vehicle does not reach the goal or does not hit the wall.}''
This explanation captures a failure case, where the ego vehicle sometimes does not reach the goal because it collides with the other vehicle, though it does successfully avoid walls.
This result suggests that our explanations can capture unlikely, but observed, behaviors from the target policy that occur due to environment stochasticity.
We also performed an ablation study, where we independently removed the extension step, expansion step, and weights in calculating KL divergence (i.e., we set $w_i = 1$ in \Cref{eqn:weight}), and found that none of the modified versions of our method successfully found the target explanation.

\Cref{tab:robot_search} shows our robot navigation results.
For the LTL target policy, our method found the target explanation in Searches 1-8.
For the non-LTL target policy, our method identified the best explanation as: ``\textit{Eventually, the robot reaches the goal or does not hit the vase. Globally, the robot does not enter a hazard}'' (Search 1).
While the task part of the explanation may seem unexpected, it is actually reasonable because the target policy did not always reach the goal (the success rate was around 60\%) and it occasionally hit the vase (since it was not penalized for doing so) --- since our method has no mechanism for ignoring predicates, the vase predicate ended up in the task clause. The constraint part of the explanation was expected, since the target policy learned to avoid hazards.
This result suggests that our method can find a reasonable explanation, even for policies not optimized for a candidate explanation.

\rowcolors{2}{gray!25}{white}
\begin{table*}[tb]
    \centering
    \medskip
    \caption{Robot navigation results for LTL (left) \& non-LTL (right) target policies.}
    \begin{tabular}{c|lcc|lcc}
         Search & Robot explanations    & wKL div. [-]  & \begin{tabular}{c} Searched\\ specs [\%] \end{tabular} & Robot explanations    & wKL div. [-]  & \begin{tabular}{c} Searched\\ specs [\%] \end{tabular} \\
         \hline
         1  & $\pmb{\mathcal{F}(\psi_\mathrm{gl}) \wedge \mathcal{G}(\neg\psi_\mathrm{hz} \wedge \neg\psi_\mathrm{vs})}$   & $\mathbf{0.00}$                & \textbf{21.8}  & $\pmb{\mathcal{F}(\psi_\mathrm{gl} \vee \neg\psi_\mathrm{vs}) \wedge \mathcal{G}(\neg\psi_\mathrm{hz})}$    &  $\mathbf{3.0501 \times 10^{-4}}$  & \textbf{39.6}\\
         2  & $\pmb{\mathcal{F}(\psi_\mathrm{gl}) \wedge \mathcal{G}(\neg\psi_\mathrm{hz} \wedge \neg\psi_\mathrm{vs})}$   & $\mathbf{0.00}$                & \textbf{30.2}  & $\mathcal{F}(\neg\psi_\mathrm{gl}) \wedge \mathcal{G}(\neg\psi_\mathrm{hz} \vee \neg\psi_\mathrm{vs})$  &  $3.0504 \times 10^{-4}$  & 29.2 \\
         3  & $\pmb{\mathcal{F}(\psi_\mathrm{gl}) \wedge \mathcal{G}(\neg\psi_\mathrm{hz} \wedge \neg\psi_\mathrm{vs})}$   & $\mathbf{0.00}$                 & \textbf{26.0}  & $\mathcal{F}(\neg\psi_\mathrm{gl}) \wedge \mathcal{G}(\neg\psi_\mathrm{hz} \vee \neg\psi_\mathrm{vs})$  &  $3.0504 \times 10^{-4}$  & 27.1\\
         4  & $\pmb{\mathcal{F}(\psi_\mathrm{gl}) \wedge \mathcal{G}(\neg\psi_\mathrm{hz} \wedge \neg\psi_\mathrm{vs})}$      & $\mathbf{0.00}$                  &\textbf{29.2}  & $\mathcal{F}(\neg\psi_\mathrm{gl}) \wedge \mathcal{G}(\neg\psi_\mathrm{hz} \vee \neg\psi_\mathrm{vs})$  &  $3.0504 \times 10^{-4}$  & 27.1\\
         5  & $\pmb{\mathcal{F}(\psi_\mathrm{gl}) \wedge \mathcal{G}(\neg\psi_\mathrm{hz} \wedge \neg\psi_\mathrm{vs})}$      & $\mathbf{0.00}$                  & \textbf{21.9}  & $\mathcal{F}(\neg\psi_\mathrm{gl}) \wedge \mathcal{G}(\neg\psi_\mathrm{hz} \vee \neg\psi_\mathrm{vs})$  &  $3.0504 \times 10^{-4}$  & 29.2\\
         6  & $\pmb{\mathcal{F}(\psi_\mathrm{gl}) \wedge \mathcal{G}(\neg\psi_\mathrm{hz} \wedge \neg\psi_\mathrm{vs})}$      & $\mathbf{0.00}$                  & \textbf{24.0}  & $\mathcal{F}(\neg\psi_\mathrm{gl}) \wedge \mathcal{G}(\neg\psi_\mathrm{hz} \vee \neg\psi_\mathrm{vs})$  &  $3.0504 \times 10^{-4}$  & 33.3\\
         7  & $\mathcal{F}(\neg\psi_\mathrm{vs}) \wedge \mathcal{G}(\psi_\mathrm{gl} \vee \neg\psi_\mathrm{hz})$     & $2.64 \times 10^{-4}$ & 25.0  & $\mathcal{F}(\psi_\mathrm{gl} \vee \psi_\mathrm{hz}) \wedge \mathcal{G}(\neg\psi_\mathrm{vs})$  &  $3.0508 \times 10^{-4}$    & 29.2\\
         8  & $\mathcal{F}(\neg\psi_\mathrm{vs}) \wedge \mathcal{G}(\psi_\mathrm{gl} \vee \neg\psi_\mathrm{hz})$      & $2.64 \times 10^{-4}$ & 27.1  & $\mathcal{F}(\psi_\mathrm{gl} \vee \psi_\mathrm{hz}) \wedge \mathcal{G}(\neg\psi_\mathrm{vs})$  &  $3.0508 \times 10^{-4}$    & 21.9\\
         \hiderowcolors
    \end{tabular}
    \label{tab:robot_search}
\end{table*}

\section{Conclusions}
\label{sec:conclusions}

This paper presents a method for generating an explanation of a given RL policy using a connected class of LTL formulae. We employ a local-search algorithm that identifies the explanation which produces a policy that best matches the target policy, based on a weighted KL divergence metric. We verified our approach in three simulated environments. Limitations include computationally complexity (we scale exponentially with the number of predicates and require an RL optimization for searched candidates), requiring user-defined predicates, interpreting inferred LTL formulas, and assuming stationary, near-optimal policies. Potential improvements are automating predicate selection, translating formulae to natural language, integrating into RL algorithms for explainability-by-design, scaling through neural network LTL representations, reducing computation through transfer learning or optimization landscape exploration, and bounding the number of required searches.

\printbibliography

@article{arora2021a,
  title = {A Survey of Inverse Reinforcement Learning: {{Challenges}}, Methods and Progress},
  author = {Arora, Saurabh and Doshi, Prashant},
  year = {2021},
  journal = {Artificial Intelligence},
  volume = {297},
  eprint = {1806.06877},
  pages = {103500},
  publisher = {Elsevier B.V.},
  issn = {00043702},
  doi = {10.1016/j.artint.2021.103500},
  url = {https://doi.org/10.1016/j.artint.2021.103500},
  archiveprefix = {arXiv}
}

@inproceedings{bewley2021,
  title = {{{TripleTree}}: {{A Versatile Interpretable Representation}} of {{Black Box Agents}} and Their {{Environments}}},
  shorttitle = {{{TripleTree}}},
  booktitle = {Proceedings of the {{AAAI Conference}} on {{Artificial Intelligence}}},
  author = {Bewley, Tom and Lawry, Jonathan},
  year = {2021},
  month = may,
  volume = {35},
  pages = {11415--11422},
  doi = {10.1609/aaai.v35i13.17360},
  url = {https://ojs.aaai.org/index.php/AAAI/article/view/17360},
  urldate = {2024-10-16},
  langid = {english}
}

@inproceedings{lundberg2017,
  title = {A {{Unified Approach}} to {{Interpreting Model Predictions}}},
  booktitle = {31st {{Conference}} on {{Neural Information Processing Systems}} ({{NIPS}} 2017)},
  author = {Lundberg, Scott M and Lee, Su-In},
  year = {2017},
  langid = {english}
}

@inproceedings{madumal2020,
  title = {Explainable Reinforcement Learning through a Causal Lens},
  booktitle = {{{AAAI}} 2020 - 34th {{AAAI Conference}} on {{Artificial Intelligence}}},
  author = {Madumal, Prashan and Miller, Tim and Sonenberg, Liz and Vetere, Frank},
  year = {2020},
  eprint = {1905.10958},
  pages = {2493--2500},
  issn = {2159-5399},
  doi = {10.1609/aaai.v34i03.5631},
  archiveprefix = {arXiv},
  isbn = {978-1-57735-835-0}
}

@inproceedings{ribeiro2016,
  title = {"{{Why Should I Trust You}}?": {{Explaining}} the {{Predictions}} of {{Any Classifier}}},
  shorttitle = {"{{Why Should I Trust You}}?},
  booktitle = {Proceedings of the 22nd {{ACM SIGKDD International Conference}} on {{Knowledge Discovery}} and {{Data Mining}}},
  author = {Ribeiro, Marco Tulio and Singh, Sameer and Guestrin, Carlos},
  year = {2016},
  month = aug,
  pages = {1135--1144},
  publisher = {ACM},
  address = {San Francisco California USA},
  doi = {10.1145/2939672.2939778},
  url = {https://dl.acm.org/doi/10.1145/2939672.2939778},
  urldate = {2024-11-20},
  isbn = {978-1-4503-4232-2},
  langid = {english}
}

@article{silver_general_2018,
	title = {A general reinforcement learning algorithm that masters chess, shogi, and Go through self-play},
	volume = {362},
	url = {https://www.science.org/doi/10.1126/science.aar6404},
	doi = {10.1126/science.aar6404},
	pages = {1140--1144},
	number = {6419},
	journaltitle = {Science},
	author = {Silver, David and Hubert, Thomas and Schrittwieser, Julian and Antonoglou, Ioannis and Lai, Matthew and Guez, Arthur and Lanctot, Marc and Sifre, Laurent and Kumaran, Dharshan and Graepel, Thore and Lillicrap, Timothy and Simonyan, Karen and Hassabis, Demis},
	urldate = {2024-09-09},
	date = {2018-12-07},
	note = {Publisher: Amer. Association for the Advancement of Science},
}

@article{heuillet_explainability_2021,
	title = {Explainability in deep reinforcement learning},
	volume = {214},
	issn = {0950-7051},
	url = {https://www.sciencedirect.com/science/article/pii/S0950705120308145},
	doi = {10.1016/j.knosys.2020.106685},
	pages = {106685},
	journaltitle = {Knowledge-Based Syst.},
	author = {Heuillet, Alexandre and Couthouis, Fabien and Díaz-Rodríguez, Natalia},
	urldate = {2024-09-09},
	date = {2021-02-28},
}

@article{bartocci_survey_2022,
	title = {Survey on mining signal temporal logic specifications},
	volume = {289},
	issn = {0890-5401},
	url = {https://www.sciencedirect.com/science/article/pii/S0890540122001122},
	doi = {10.1016/j.ic.2022.104957},
	pages = {104957},
	journaltitle = {Inf. and Computation},
	author = {Bartocci, Ezio and Mateis, Cristinel and Nesterini, Eleonora and Nickovic, Dejan},
	urldate = {2024-09-09},
	date = {2022-11-01},
}

@article{li_formal_2019,
	title = {A formal methods approach to interpretable reinforcement learning for robotic planning},
	volume = {4},
	url = {https://www.science.org/doi/10.1126/scirobotics.aay6276},
	doi = {10.1126/scirobotics.aay6276},
	pages = {eaay6276},
	number = {37},
	journaltitle = {Science Robot.},
	author = {Li, Xiao and Serlin, Zachary and Yang, Guang and Belta, Calin},
	urldate = {2024-09-09},
	date = {2019-12-18},
	note = {Publisher: Amer. Association for the Advancement of Science},
}

@article{bombara_offline_2021,
	title = {Offline and Online Learning of Signal Temporal Logic Formulae Using Decision Trees},
	volume = {5},
	issn = {2378-962X},
	url = {https://dl.acm.org/doi/10.1145/3433994},
	doi = {10.1145/3433994},
	pages = {22:1--22:23},
	number = {3},
	journaltitle = {{ACM} Trans. Cyber-Phys. Syst.},
	author = {Bombara, Giuseppe and Belta, Calin},
	urldate = {2024-09-09},
	date = {2021},
}

@inproceedings{gaglione_learning_2021,
	location = {Cham},
	title = {Learning Linear Temporal Properties from Noisy Data: A {MaxSAT}-Based Approach},
	isbn = {978-3-030-88885-5},
	doi = {10.1007/978-3-030-88885-5_6},
	shorttitle = {Learning Linear Temporal Properties from Noisy Data},
	pages = {74--90},
	booktitle = {Automated Technology for Verification and Analysis},
	publisher = {Springer Int. Publishing},
	author = {Gaglione, Jean-Raphaël and Neider, Daniel and Roy, Rajarshi and Topcu, Ufuk and Xu, Zhe},
	editor = {Hou, Zhe and Ganesh, Vijay},
	date = {2021},
	langid = {english},
}

@inproceedings{li_learning_2023,
	title = {Learning Signal Temporal Logic through Neural Network for Interpretable Classification},
	url = {https://ieeexplore.ieee.org/document/10156357},
	doi = {10.23919/ACC55779.2023.10156357},
	eventtitle = {2023 Amer. Control Conf. ({ACC})},
	pages = {1907--1914},
	booktitle = {2023 Amer. Control Conf. ({ACC})},
	author = {Li, Danyang and Cai, Mingyu and Vasile, Cristian-Ioan and Tron, Roberto},
	urldate = {2024-09-09},
	date = {2023-05},
	note = {{ISSN}: 2378-5861},
}

@inproceedings{jha_telex_2017,
	location = {Cham},
	title = {{TeLEx}: Passive {STL} Learning Using Only Positive Examples},
	isbn = {978-3-319-67531-2},
	doi = {10.1007/978-3-319-67531-2_13},
	shorttitle = {{TeLEx}},
	pages = {208--224},
	booktitle = {Runtime Verification},
	publisher = {Springer Int. Publishing},
	author = {Jha, Susmit and Tiwari, Ashish and Seshia, Sanjit A. and Sahai, Tuhin and Shankar, Natarajan},
	editor = {Lahiri, Shuvendu and Reger, Giles},
	date = {2017},
	langid = {english},
}

@article{jha_telex_2019,
	title = {{TeLEx}: learning signal temporal logic from positive examples using tightness metric},
	volume = {54},
	issn = {1572-8102},
	url = {https://doi.org/10.1007/s10703-019-00332-1},
	doi = {10.1007/s10703-019-00332-1},
	shorttitle = {{TeLEx}},
	pages = {364--387},
	number = {3},
	journaltitle = {Formal Methods Syst. Des.},
	author = {Jha, Susmit and Tiwari, Ashish and Seshia, Sanjit A. and Sahai, Tuhin and Shankar, Natarajan},
	urldate = {2024-09-09},
	date = {2019-11-01},
	langid = {english},
}

@article{sutton_between_1999,
	title = {Between {MDPs} and semi-{MDPs}: A framework for temporal abstraction in reinforcement learning},
	volume = {112},
	issn = {0004-3702},
	url = {https://www.sciencedirect.com/science/article/pii/S0004370299000521},
	doi = {10.1016/S0004-3702(99)00052-1},
	shorttitle = {Between {MDPs} and semi-{MDPs}},
	pages = {181--211},
	number = {1},
	journaltitle = {Artificial Intell.},
	author = {Sutton, Richard S. and Precup, Doina and Singh, Satinder},
	urldate = {2024-09-09},
	date = {1999-08-01},
}

@inproceedings{schneider_improving_2001,
	location = {Berlin, Heidelberg},
	title = {Improving Automata Generation for Linear Temporal Logic by Considering the Automaton Hierarchy},
	isbn = {978-3-540-42957-9},
	series = {{LPAR} '01},
	pages = {39--54},
	booktitle = {Proc. Artificial Intell. Logic Program.},
	publisher = {Springer-Verlag},
	author = {Schneider, Klaus},
	urldate = {2024-09-09},
	date = {2001},
}

@misc{schulman_proximal_2017,
	title = {Proximal Policy Optimization Algorithms},
	url = {http://arxiv.org/abs/1707.06347},
	doi = {10.48550/arXiv.1707.06347},
	number = {{arXiv}:1707.06347},
	publisher = {{arXiv}},
	author = {Schulman, John and Wolski, Filip and Dhariwal, Prafulla and Radford, Alec and Klimov, Oleg},
	urldate = {2024-09-09},
	date = {2017-08-28},
	eprinttype = {arxiv},
	eprint = {1707.06347 [cs]},
}

@inproceedings{haarnoja_soft_2018,
	title = {Soft Actor-Critic: Off-Policy Maximum Entropy Deep Reinforcement Learning with a Stochastic Actor},
	url = {https://proceedings.mlr.press/v80/haarnoja18b.html},
	shorttitle = {Soft Actor-Critic},
	eventtitle = {Int. Conf. Mach. Learn.},
	pages = {1861--1870},
	booktitle = {Proc. of the 35th Int. Conf. Mach. Learn.},
	publisher = {{PMLR}},
	author = {Haarnoja, Tuomas and Zhou, Aurick and Abbeel, Pieter and Levine, Sergey},
	urldate = {2024-09-09},
	date = {2018-07-03},
	langid = {english},
	note = {{ISSN}: 2640-3498},
}

@inproceedings{andrychowicz_hindsight_2017,
	title = {Hindsight Experience Replay},
	volume = {30},
	url = {https://papers.nips.cc/paper/2017/hash/453fadbd8a1a3af50a9df4df899537b5-Abstract.html},
	booktitle = {Advances in Neural Inf. Process. Syst.},
	publisher = {Curran Associates, Inc.},
	author = {Andrychowicz, Marcin and Wolski, Filip and Ray, Alex and Schneider, Jonas and Fong, Rachel and Welinder, Peter and {McGrew}, Bob and Tobin, Josh and Pieter Abbeel, {OpenAI} and Zaremba, Wojciech},
	urldate = {2024-09-09},
	date = {2017},
}

@misc{highway-env,
  author = {Leurent, Edouard},
  title = {An Environment for Autonomous Driving Decision-Making},
  year = {2018},
  publisher = {GitHub},
  journal = {GitHub repository},
  howpublished = {https://github.com/eleurent/highway-env},
}

@inproceedings{yan_stone_2021,
	title = {{STONE}: Signal Temporal Logic Neural Network for Time Series Classification},
	url = {https://ieeexplore.ieee.org/document/9679844},
	doi = {10.1109/ICDMW53433.2021.00101},
	shorttitle = {{STONE}},
	eventtitle = {2021 Int. Conf. Data Mining Workshops ({ICDMW})},
	pages = {778--787},
	booktitle = {2021 Int. Conf. Data Mining Workshops ({ICDMW})},
	author = {Yan, Ruixuan and Julius, Agung and Chang, Maria and Fokoue, Achille and Ma, Tengfei and Uceda-Sosa, Rosario},
	urldate = {2024-09-09},
	date = {2021-02},
	note = {{ISSN}: 2375-9259},
}

@article{karagulle_safe_2024,
	title = {A Safe Preference Learning Approach for Personalization With Applications to Autonomous Vehicles},
	volume = {9},
	issn = {2377-3766},
	url = {https://ieeexplore.ieee.org/abstract/document/10465615},
	doi = {10.1109/LRA.2024.3375626},
	number = {5},
	urldate = {2024-08-16},
	journal = {IEEE Robot. Automat. Lett.},
	author = {Karagulle, Ruya and Aréchiga, Nikos and Best, Andrew and DeCastro, Jonathan and Ozay, Necmiye},
	month = may,
	year = {2024},
	note = {Conf. Name: IEEE Robot. Automat. Lett.},
	pages = {4226--4233},
}

@article{kong_temporal_2017,
	title = {Temporal Logics for Learning and Detection of Anomalous Behavior},
	volume = {62},
	issn = {1558-2523},
	url = {https://ieeexplore.ieee.org/abstract/document/7500142},
	doi = {10.1109/TAC.2016.2585083},
	pages = {1210--1222},
	number = {3},
	journaltitle = {{IEEE} Trans. Automat. Control},
	author = {Kong, Zhaodan and Jones, Austin and Belta, Calin},
	urldate = {2024-09-11},
	date = {2017-03},
	note = {Conference Name: {IEEE} Transactions on Automatic Control},
}

@article{ju_transferring_2022,
	title = {Transferring policy of deep reinforcement learning from simulation to reality for robotics},
	volume = {4},
	rights = {2022 Springer Nature Limited},
	issn = {2522-5839},
	url = {https://www.nature.com/articles/s42256-022-00573-6},
	doi = {10.1038/s42256-022-00573-6},
	pages = {1077--1087},
	number = {12},
	journaltitle = {Nat. Mach. Intell.},
	author = {Ju, Hao and Juan, Rongshun and Gomez, Randy and Nakamura, Keisuke and Li, Guangliang},
	urldate = {2024-09-13},
	date = {2022-12},
	langid = {english},
	note = {Publisher: Nature Publishing Group},
}

@article{stable-baselines3,
  author  = {Antonin Raffin and Ashley Hill and Adam Gleave and Anssi Kanervisto and Maximilian Ernestus and Noah Dormann},
  title   = {Stable-Baselines3: Reliable Reinforcement Learning Implementations},
  journal = {J. Mach. Learn. Res.},
  year    = {2021},
  volume  = {22},
  number  = {268},
  pages   = {1-8},
  url     = {http://jmlr.org/papers/v22/20-1364.html}
}

@misc{rl-zoo3,
  author = {Raffin, Antonin},
  title = {RL Baselines3 Zoo},
  year = {2020},
  publisher = {GitHub},
  journal = {GitHub repository},
  howpublished = {\url{https://github.com/DLR-RM/rl-baselines3-zoo}},
}

@inproceedings{ji2023safety,
  title={Safety Gymnasium: A Unified Safe Reinforcement Learning Benchmark},
  author={Jiaming Ji and Borong Zhang and Jiayi Zhou and Xuehai Pan and Weidong Huang and Ruiyang Sun and Yiran Geng and Yifan Zhong and Josef Dai and Yaodong Yang},
  booktitle={37th Conf. Neural Info. Process. Syst. Datasets and Benchmarks Track},
  year={2023},
  url={https://openreview.net/forum?id=WZmlxIuIGR}
}

@inproceedings{guo2021bb,
  title = {{{EDGE}}: {{Explaining Deep Reinforcement Learning Policies}}},
  booktitle = {35th {{Conf.}} on {{Neural Inf. Process. Syst.}} ({{NeurIPS}} 2021)},
  author = {Guo, Wenbo and Wu, Xian and Khan, Usmann and Xing, Xinyu},
  year = {2021},
  langid = {english}
}

\vfill

\end{document}